\documentclass[conference]{IEEEtran}
\usepackage{cite}
\usepackage{amsmath,amssymb}
\usepackage{graphicx}
\usepackage{hyperref}
\usepackage{tabularx}
\usepackage{adjustbox}
\usepackage{caption}       % fine-grained caption control
\usepackage{booktabs}

\title{Scaling Arabic Medical Chatbots Using Synthetic Data: Enhancing Generative AI with Synthetic Patient Records}

\author{
\IEEEauthorblockN{{\large Abdulrahman Allam\IEEEauthorrefmark{1},
Seif Ahmed\IEEEauthorrefmark{1},
Ali Hamdi\IEEEauthorrefmark{1},
Khaled Shaban\IEEEauthorrefmark{2}}}

\IEEEauthorblockA{\IEEEauthorrefmark{1}\large\textit{Dept. of Computer Science, MSA University, Giza, Egypt} \\
\{abdulrahman.atif, seifeldein.ahmed, ahamdi\}@msa.edu.eg}

\IEEEauthorblockA{\IEEEauthorrefmark{2}\large\textit{Dept. of Computer Science, Qatar University, Doha, Qatar} \\
khaled.shaban@qu.edu.qa}
}

\begin{document}
\maketitle

\begin{abstract}
The development of medical chatbots in Arabic is significantly constrained by the scarcity of large-scale, high-quality annotated datasets. While prior efforts compiled a dataset of 20,000 Arabic patient–doctor interactions from social media to fine-tune large language models (LLMs), model scalability and generalization remained limited. In this study, we propose a scalable synthetic data augmentation strategy to expand the training corpus to 100,000 records. Using advanced generative AI systems—ChatGPT-4o and Gemini 2.5 Pro—we generated 80,000 contextually relevant and medically coherent synthetic question–answer pairs grounded in the structure of the original dataset. These synthetic samples were semantically filtered, manually validated, and integrated into the training pipeline. We fine-tuned five LLMs, including Mistral-7B and AraGPT2, and evaluated their performance using BERTScore metrics and expert-driven qualitative assessments. To further analyze the effectiveness of synthetic sources, we conducted an ablation study comparing ChatGPT-4o and Gemini-generated data independently. The results showed that ChatGPT-4o data consistently led to higher F1-scores and fewer hallucinations across all models. Overall, our findings demonstrate the viability of synthetic augmentation as a practical solution for enhancing domain-specific language models in low-resource medical NLP, paving the way for more inclusive, scalable, and accurate Arabic healthcare chatbot systems.
\end{abstract}


\section{Introduction}

The growing global demand for healthcare services, along with limited infrastructure and increased patient expectations, has heightened the need for novel technological solutions to improve disease detection, emergency response, and overall resource allocation. \cite{detmer1997computer,brown1984concept,rashid2025ai}. Many portions of the world, particularly Arabic-speaking countries, continue to struggle to deliver timely, accurate, and context-aware help due to both infrastructure and linguistic problems. Conventional solutions frequently rely on rule-based or classical machine learning (ML) models, which are unable to deal with the informal, unstructured, and dialectally heterogeneous nature of real-world medical questions. \cite{kuhl2022artificial,priyanka2024hospital}.

Recent breakthroughs in natural language processing (NLP) and generative artificial intelligence (AI) have resulted in large language models (LLMs) that can generate coherent, contextually rich responses. These approaches, when correctly applied to the medical domain, provide transformative possibilities for healthcare delivery in under-resourced languages like Arabic. \cite{nazi2024large,peng2023study}. However, fine-tuning LLMs for particular tasks needs access to huge volumes of domain-specific data, which is typically scarce in the Arabic medical setting. \cite{abdelhay2023deep,al2020nabiha}.

To address this data scarcity, we compiled a curated dataset of 20,000 real-world patient-doctor interactions from Arabic-language social media, capturing informal speech, dialectal variance, and genuine medical concerns. This dataset served as a foundation for fine-tuning models like Mistral-7B, LLaMA-2-7B, and AraGPT2, yielding promising results in generating context-aware medical responses in Arabic.

Despite this progress, establishing generalization and resilience in LLMs requires increasingly larger and more diverse training datasets. Unfortunately, manually increasing real-world Arabic medical databases is time-consuming and fraught with ethical, legal, and privacy considerations. To alleviate this barrier, we use a synthetic data creation strategy, which is increasingly acknowledged for its capacity to grow datasets while protecting data utility and privacy. \cite{figueira2022survey,lu2023machine,murtaza2023synthetic,dahmen2019synsys,dankar2021fake}.

In this study, we use synthetic patient-doctor interactions to increase the size of our Arabic medical dataset by fivefold, from 20,000 to 100,000 entries. We used cutting-edge generative AI systems, notably ChatGPT-4o and Gemini 2.5 Pro, to create 80,000 new samples based on the language and structural trends in our original dataset. These synthetic recordings were meticulously chosen to ensure contextual relevance, dialectal accuracy, and clinical plausibility. We then re-fine-tuned our LLMs on the additional dataset and conducted a thorough evaluation to compare performance improvements.

Our findings indicate that training with synthetic data considerably improves model adaptability, lowers overfitting, and increases output diversity across various evaluation criteria. Furthermore, this study provides a consistent framework for using synthetic data to improve generative AI applications in low-resource healthcare settings.

The rest of the paper is organized as follows: Section \ref{related-work} reviews existing literature on LLMs in healthcare and synthetic data generation. Section \ref{problem} shows the challenges and obstacles. Section \ref{proposed-model} details the dataset construction, synthetic data generation pipeline, and model training procedures. Section \ref{experiments} presents the experimental design and evaluation metrics. Section \ref{results} analyzes the results, Section \ref{ablation} explores the ablation study, and Section \ref{conclusion} for conclusion.

\begin{figure*}[h] % h = here, t = top, b = bottom, p = page
    \centering
    \includegraphics[width=1\textwidth]{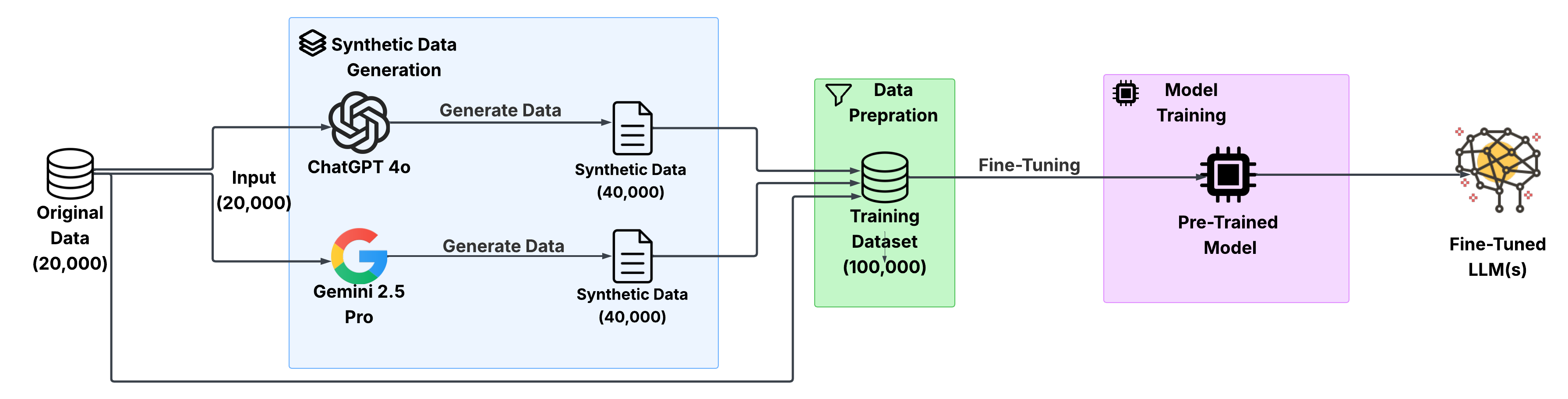}
    \caption{Overview of the proposed synthetic data augmentation and model fine-tuning pipeline.}
    \label{fig:ffff}
\end{figure*}

\section{Related Work}
\label{related-work}

The use of large language models (LLMs) into healthcare systems has resulted in major improvements to clinical decision support, patient engagement, and automated medical recording. These algorithms, trained on large biological corpora, can recognize medical terminology, generate patient summaries, and even offer treatment regimens. \cite{peng2023study, zhang2023generative, nazi2024large}. In English-speaking contexts, LLMs have been fine-tuned with high-quality structured datasets such as electronic health records (EHRs) and clinical notes. However, in Arabic-language healthcare settings, the scarcity of annotated datasets, along with significant dialectal variety and informal language usage, creates particular obstacles. \cite{abdelhay2023deep}.

Traditionally, Arabic medical chatbots have used rule-based systems or shallow machine learning models that rely on predetermined templates or manually vetted responses. \cite{rekik2023, al2020nabiha}. While these approaches provide explainability and control, they do not generalize to unstructured, colloquial, or dialect-rich inputs. This issue is particularly noticeable in online patient-doctor communications, which frequently lack conventional syntax, punctuation, and medical specificity. \cite{anwar2025towards, priyanka2024hospital}. Systems like PharmaGo \cite{gamage2021} and NABiha \cite{al2020nabiha} proved that, while rule-based models can address specific medical use cases, they do not scale to larger real-world contexts or adapt effectively to informal patient language.

To solve these problems, subsequent works have examined the use of generative AI and fine-tuned LLMs like GPT-2, LLaMA-2, and BLOOM in Arabic NLP. \cite{kuhl2022artificial, abdellaif2024lmrpa, hamdi2024llm}. When these models are trained with domain-specific data, they can generate grammatically consistent and context-sensitive medical output. For instance, Abdelhay et al. \cite{abdelhay2023deep} demonstrated the potential of Arabic LLMs fine-tuned on real patient-doctor dialogues collected from Egyptian medical forums and social media.

Despite these developments, the size and diversity of training data continue to be key barriers for the effective deployment of Arabic LLMs. Manual data gathering in the medical area is time-consuming and constrained by ethical considerations and data protection rules. As a result, researchers have increasingly turned to synthetic data production as a feasible approach to enhance constrained real-world information. \cite{figueira2022survey, murtaza2023synthetic, dahmen2019synsys}. Synthetic data has shown promise in replicating uncommon medical events, protecting patient privacy, and increasing model generalization. \cite{dankar2021fake, aldea2024using}. While these strategies are common in English-language healthcare AI, synthetic augmentation in Arabic-language NLP—especially in the medical domain—remains underexplored.

Notably, prior work such as RIRO \cite{hamdi2024riro} and ERPA \cite{abdellaif2024erpa} has demonstrated the usefulness of LLMs in low-resource and document-processing settings. These studies demonstrate the ability of huge models to operate in data-scarce environments by reshaping inputs and refining output. Similarly, techniques like data synthesis, transfer learning, and low-rank adaptation (LoRA) are being used to lessen reliance on big annotated datasets.

Our current work builds upon this growing line of research by designing a structured pipeline for Arabic medical synthetic data generation. Using ChatGPT-4o and Gemini 2.5 Pro, we generate 80,000 synthetic patient–doctor interactions, carefully aligned with linguistic and contextual patterns observed in real-world clinical settings. We demonstrate that augmenting the training corpus with synthetic samples improves model fluency, robustness to noisy input, and adaptability across medical subdomains. This contribution aligns with a growing body of research advocating for synthetic data as an enabler of scalable and ethical AI development in healthcare, particularly for underrepresented languages.

\section{Problem Formulation}
\label{problem}

Despite enormous progress in large language models (LLMs), their efficacy in specialized domains such as healthcare is still dependent on availability to large-scale, high-quality, domain-specific data. This dependency creates numerous significant obstacles in Arabic medical NLP, limiting the development and implementation of accurate, scalable, and generalizable chatbot systems.

\subsection{Limited Access to High-Quality Arabic Medical Data}

The great majority of publicly available medical datasets are in English and originate from Western healthcare infrastructures. In contrast, Arabic-language resources are few, particularly those having structured patient-doctor conversations. Available Arabic datasets are frequently domain-general, lack standardization, or contain clinical content that is inadequately annotated or unrepresentative of informal patient conversation. Our previously curated collection of 20,000 real-world interactions derived from social media posts represented a critical first step toward solving this paucity. It recorded valuable language diversity, dialectal idioms, and accurate symptom descriptions. However, the dataset was still insufficient in size to adequately allow the training of high-capacity generative models capable of handling unusual situations, unclear inputs, or edge-case scenarios.

\subsection{High Cost and Complexity of Manual Data Collection}

The manual collection of Arabic medical data presents logistical, ethical, and technical obstacles. Access to clinical records is limited due to privacy regulations and ethical review limits. Even in publicly available contexts like forums and social media, obtaining valuable medical data needs significant effort to find relevant information, confirm medical context, and prepare for model preparation. Dialectal variety within Arabic-speaking regions adds another degree of intricacy, as the same symptoms can be expressed differently depending on the speaker's background or education level. Furthermore, annotation for medical NLP applications frequently necessitates input from domain experts, which greatly increases time and resource requirements. Preprocessing this data—removing noise, resolving ambiguity, and normalizing syntax—is not straightforward and can have a significant impact on downstream model performance if not handled carefully.

\subsection{Underutilization of Synthetic Data in Arabic Medical NLP}

While synthetic data generation has gained traction in fields such as cybersecurity, finance, and English-language healthcare AI \cite{figueira2022survey, lu2023machine, murtaza2023synthetic}, Its potential for Arabic medical NLP is mostly untapped. Synthetic data can be used to imitate uncommon disease instances, safeguard patient privacy, and provide linguistic diversity to model training corpora. \cite{dahmen2019synsys, dankar2021fake}.However, there is no empirical evidence in Arabic NLP—particularly in healthcare—to assess the quality, coherence, and utility of machine-generated patient-doctor interactions. Concerns remain concerning whether synthetic examples may maintain contextual integrity, reflect true medical thinking, and generalize well across unseen user inputs. The true benefit of synthetic augmentation in this sector remains unknown due to a lack of systematic validation.

\subsection{Challenge of Ensuring Data Quality for Training}

Even when synthetic data is available, determining its suitability for model training is a substantial challenge. Generative outputs must be filtered and modified to eliminate factual errors, language incoherence, and misleading medical advice. Training LLMs on badly produced synthetic data can lead to misinformation spread, poor performance, and biased results. Thus, a major problem is not just creating synthetic data, but also evaluating and curating it to ensure that it meets real-world language and clinical criteria. Effective synthetic augmentation necessitates a delicate balance of variation and plausibility.

\subsection{Research Objective}

This study looks on the effects of large-scale synthetic data augmentation on the performance of Arabic medical LLMs. We investigate whether synthetic records created from fundamental real-world data utilizing advanced LLMs such as ChatGPT-4o and Gemini 2.5 Pro can improve the fluency, contextual comprehension, and generalization capabilities of medical chatbot systems. By increasing our dataset from 20,000 to 100,000 records, we aim to:

\begin{itemize}
    \item Evaluate the semantic diversity and medical relevance of synthetic Arabic patient–doctor dialogues.
    \item Retrain and fine-tune existing generative models (e.g., Mistral-7B, LLaMA-2-7B, AraGPT2) using both real and synthetic data.
    \item Assess model performance across BERTScore metrics (precision, recall, F1-score) and qualitative measures such as coherence and informativeness.
    \item Examine the role of synthetic data in reducing overfitting, enhancing robustness to noisy inputs, and supporting rare condition inference.
\end{itemize}

Our central hypothesis is that high-quality synthetic data generated with dialect, context, and medical accuracy can significantly expand the training corpus for Arabic medical NLP and facilitate the development of more inclusive, scalable, and effective AI-driven healthcare solutions in low-resource settings.

Our key premise is that high-quality synthetic data can compensate for the shortage of real-world Arabic medical records, resulting in more scalable and effective chatbot systems for healthcare in low-resource settings.

\section{Proposed Model}
\label{proposed-model}
To solve the constraints of data scarcity in Arabic medical NLP and improve the generalization capabilities of large language models (LLMs), we offer a two-stage framework: (1) large-scale synthetic data production with curated seed data and cutting-edge generative models, and (2) retraining numerous LLMs on the enhanced dataset. This pipeline enables us to simulate various, medically consistent Arabic discussions and add them into the model training process. An overview of the system architecture is displayed in Fig.~\ref{fig:ffff}.

\subsection{Synthetic Data Generation Pipeline}

The core innovation of this work focuses on expansion the original 20,000-record dataset to a total of 100,000 records. To achieve this, we generated 80,000 synthetic patient–doctor interactions using two advanced generative AI platforms: \textbf{ChatGPT-4o} by OpenAI and \textbf{Gemini 2.5 Pro} by Google.

\subsubsection{Generative Models Used}

\textbf{ChatGPT-4o} is a multimodal LLM optimized for natural, coherent, and context-sensitive generation across various domains, including healthcare. Its ability to understand informal dialectal Arabic and respond with medically plausible advice made it particularly effective in generating synthetic question-answer (QA) pairs. We used carefully structured prompts that simulated real-world patient complaints, including varying tone, severity of symptoms, and demographics of users.

\textbf{Gemini 2.5 Pro} Gemini 2.5 Pro is the most recent edition of Google's generative AI model suite, and it is notable for improving factual grounding and context retention. Gemini was suggested utilizing templates extracted from clinical publications as well as our previously curated dataset. This allowed it to generate synthetic patient complaints and doctor responses in order to determine patient profiles and complaint categories.

\subsubsection{Data Generation Process}

\begin{figure}[ht]
    \centering
    \includegraphics[width=0.58\textwidth]{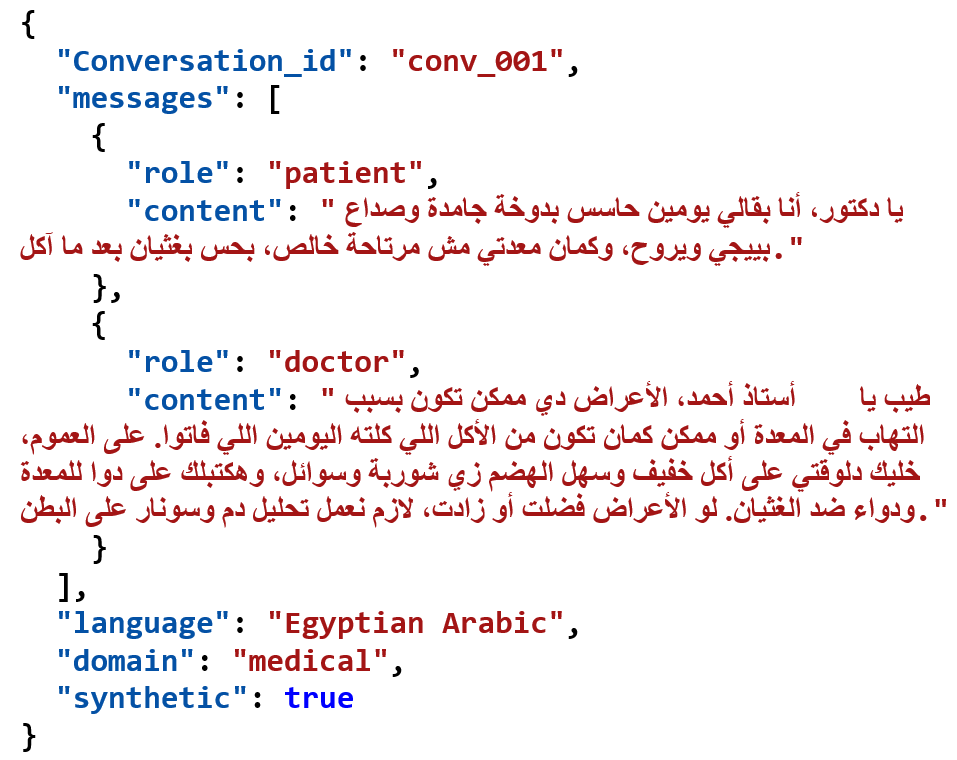}
    \caption{A sample from ChatGPT 4o Synthetic Data.}
    \label{fig:system}
\end{figure}

The generation process was guided by the original 20,000 interactions to ensure alignment with realistic structures and vocabulary. The synthetic data generation pipeline includes the following stages:

\begin{itemize}
    \item \textbf{Prompt Engineering:} A set of context-specific prompt templates was designed to simulate a wide variety of patient complaints (e.g., gastrointestinal issues, respiratory symptoms, chronic diseases) and doctor responses. Prompts were dynamically varied to reflect informal phrasing, dialectical nuances, and topic diversity.
    
    \item \textbf{Model Sampling:}ChatGPT-4o was used to produce 40,000 QA pairs, with Gemini 2.5 Pro adding another 40,000 pairs. To ensure quality, the outputs were recorded via an automated querying system .
    
    \item \textbf{Contextual Anchoring:} Each prompt was in context linked using seed examples from the original dataset to ensure coherence in medical logic and consistency with regional Arabic phrasing.
    
    \item \textbf{Semantic Filtering:} After generation, each pair of QA was passed through cosine similarity filters and BERT-based embedding analysis to assess similarity to the original data set.
    
    \item \textbf{Language Verification:} Language identification methods guaranteed that all outputs were in Arabic and devoid of code-switching and translation problems.
    
    \item \textbf{Manual Review:} A random sample of 500 synthetic QA pairs (250 from each model) was manually reviewed by native Arabic speakers with basic medical knowledge to assess grammatical fluency, medical plausibility, and coherence. Fig \ref{fig:system} represents one of the samples.
\end{itemize}

This workflow generated high-quality synthetic data of 80,000 additional samples, which were then integrated with the original 20,000 genuine encounters to create a final dataset of 100,000 patient-doctor discussion pairings.

\subsection{Data Integration and Preprocessing}

The merged dataset underwent rigorous preprocessing to ensure uniformity, linguistic integrity, and model readiness:

\begin{itemize}
    \item \textbf{Normalization:} Arabic characters were normalized to handle variant forms (e.g., Alef, Teh Marbuta), and non-linguistic symbols such as emojis, URLs, and hashtags were removed.
    
    \item \textbf{Deduplication:} To eliminate near-duplicate QA pairings, a mix of precise string matching and approximate similarity identification was performed, both within and between the real and synthetic subsets.
    
    \item \textbf{Tokenization:} Each entry was tokenized using the \texttt{AutoTokenizer} Hugging Face module, tailored to the exact architecture of the model being fine-tuned. The token sequences were padded and trimmed as needed during batch processing.
\end{itemize}

\subsection{Model Fine-Tuning}

The final 100,000-record corpus was used to fine-tune five state-of-the-art generative language models previously evaluated in our earlier work:

\begin{itemize}
    \item \textbf{AraGPT2-Base} – A GPT-2 variant pre-trained on Arabic text.
    \item \textbf{Meta-LLaMA-2-7B} – A high-performance transformer with broad contextual understanding.
    \item \textbf{Mistral-7B-Instruct-v0.2} – Instruction-tuned for question–answering tasks.
    \item \textbf{BigScience/BLOOM-560M} – A multilingual model designed for low-resource languages.
    \item \textbf{OpenAI GPT-2 Medium} – A general-purpose generative model with proven robustness.
\end{itemize}

Fine-tuning was performed using the Hugging Face \texttt{Trainer} API with \textbf{Low-Rank Adaptation (LoRA)} enabled. LoRA allowed for efficient training by reducing the number of trainable parameters, making it feasible to fine-tune large models on modest hardware resources.

\subsubsection{Training Configuration}

The models were trained under the following conditions:

\begin{itemize}
    \item \textbf{Learning Rate:} Set to 5e--5 with cosine decay and a linear warmup over the first 200 steps.
    \item \textbf{Epochs:} Ranged from 3 to 5 depending on model size and convergence behavior.
    \item \textbf{Batch Size:} Tuned between 8 and 16 based on GPU memory availability (FP16 mixed precision was used).
    \item \textbf{Evaluation Strategy:} Models were evaluated at the end of each epoch using a validation split from the combined dataset. Checkpoints were saved based on F1-score improvements.
\end{itemize}

This integrated system enabled the cost-effective growth of training data while preserving linguistic authenticity and medical coherence. The generated models demonstrated enhanced generalization, resilience to informal inquiries, and flexibility to varied medical use cases in Arabic, highlighting the practical utility of synthetic augmentation in low-resource healthcare NLP.

\section{Experimental Design}

The experimental methodology for this study was designed to systematically assess the effect of large-scale synthetic data augmentation on the performance of Arabic medical large language models (LLMs). Building on our previous work, which trained models on a real-world dataset of 20,000 records, we now use synthetic augmentation to increase the training set to 100,000 records. This section describes the setup, training parameters, assessment measures, and comparative technique used to determine the efficacy of this strategy.

\subsection{Experimental Setup}
\label{experiments}

To ensure consistency and comparability, we reused the model architectures, training frameworks, and fine-tuning procedures from our earlier study. All training was conducted using the Hugging Face \texttt{Transformers} library, combined with the \texttt{PEFT} (Parameter-Efficient Fine-Tuning) module for Low-Rank Adaptation (LoRA). This setup enabled efficient model tuning on GPU-constrained hardware while preserving the scalability required for large datasets.

Five pre-trained generative language models were selected for evaluation:

\begin{itemize}
    \item \textbf{AraGPT2-Base} – A GPT-2 variant pre-trained specifically on Arabic corpora.
    \item \textbf{Meta-LLaMA-2-7B} – A high-performance model known for strong zero-shot and few-shot capabilities.
    \item \textbf{Mistral-7B-Instruct-v0.2} – Fine-tuned for instruction-based question answering tasks.
    \item \textbf{BigScience/BLOOM-560M} – A multilingual transformer model optimized for low-resource settings.
    \item \textbf{OpenAI GPT-2 Medium (355M)} – A general-purpose generative model with moderate parameter count.
\end{itemize}

Each model was fine-tuned under two configurations:

\begin{enumerate}
    \item \textbf{Baseline:} Trained using only the original 20,000 real Arabic patient–doctor interactions.
    \item \textbf{Synthetic-Augmented:} Trained on the full 100,000-entry dataset, combining real records and 80,000 synthetic QA pairs generated using ChatGPT-4o and Gemini 2.5 Pro.
\end{enumerate}

This comparison arrangement enabled us to isolate the influence of synthetic augmentation on model performance.

\subsection{Training Configuration}

All models were fine-tuned under a unified training strategy to ensure experimental fairness. Key hyperparameters and configurations included:

\begin{itemize}
    \item \textbf{Epochs:} 3–5, with early stopping criteria based on validation loss trends. Larger models like LLaMA-2 and Mistral were trained for fewer epochs to mitigate overfitting.
    \item \textbf{Batch Size:} Set to 8, balancing GPU memory limitations and gradient stability.
    \item \textbf{Learning Rate:} Initialized at 5e--5 with a cosine decay schedule and a linear warmup over the first 200 steps to facilitate smooth convergence.
    \item \textbf{Optimizer:} AdamW was used for adaptive learning rate adjustment with weight decay regularization.
    \item \textbf{Precision:} FP16 mixed precision training was enabled to reduce memory usage and speed up training, especially for larger models.
    \item \textbf{Evaluation Frequency:} Validation was performed at the end of each epoch, with checkpoints saved based on the lowest validation loss.
    \item \textbf{Hardware:} Training was distributed across multiple NVIDIA A100 and L40 GPUs, depending on model size. Checkpointing and evaluation were GPU-accelerated to minimize runtime.
\end{itemize}

These settings were consistent across both the baseline and synthetic-augmented setups, ensuring a fair and reproducible comparison.

\subsection{Evaluation Metrics}

To evaluate model performance, we employed both automatic and human-centric metrics:

\begin{itemize}
    \item \textbf{BERTScore:} Used to compute semantic similarity between model-generated outputs and ground-truth responses. We report three submetrics: precision, recall, and F1-score, using contextual embeddings tailored for Arabic text.
    
    \item \textbf{Qualitative Analysis:} A randomly selected subset of 200 generated responses from each model was reviewed by native Arabic speakers with medical training. Criteria included:
    \begin{itemize}
        \item Grammatical fluency and coherence.
        \item Relevance to the patient complaint.
        \item Medical plausibility of the advice or diagnosis.
    \end{itemize}
    
    \item \textbf{Error Examples and Failure Case Logging:} Instances of hallucinations, repetition, or ambiguous advice were logged to analyze failure patterns across training configurations.
\end{itemize}

\subsection{Baseline and Comparison Strategy}

The primary hypothesis was whether supplementing real-world data with synthetic QA pairs enhances LLM performance on Arabic medical text creation tasks. We conducted side-by-side evaluations of baseline and synthetic-augmented models.

\begin{itemize}
    \item \textbf{Baseline Models:} Represent the original setting—models fine-tuned only on the 20,000 real patient–doctor interactions.
    
    \item \textbf{Synthetic-Augmented Models:} Models re-trained on the full 100,000-record corpus, incorporating 80,000 high-quality synthetic interactions generated using ChatGPT-4o and Gemini 2.5 Pro.
\end{itemize}

To ensure robustness, all experiments were repeated three times using different random seeds. Final metrics were averaged across runs, and where applicable, standard deviation was reported to highlight variability.

This experimental design allowed us to quantify the benefits of synthetic data augmentation and examine its impact on both model fluency and semantic accuracy in Arabic medical chatbot applications.

\begin{table*}[t]
  \centering
  
  \begin{adjustbox}{max width=\textwidth}
    \begin{tabular}{|l|c|c|c|}
      \hline
      \textbf{Model Name} & \textbf{Base Model (\%)} & \textbf{Fine-Tuned (20K)} & \textbf{Fine-Tuned (100K)} \\
      \hline
      Meta-LLaMA-2-7B              & 64.00 & 67.25 & 75.03 \\
      AraGPT2-Base (148M)          & 55.45 & 65.04 & 72.30 \\
      OpenAI GPT-2 Medium (355M)   & 53.61 & 65.07 & 69.14 \\
      Mistral-7B-Instruct-v0.2     & 65.77 & 68.50 & \textbf{81.36} \\
      BigScience/BLOOM-560M        & 50.12 & 65.60 & 68.89 \\
      \hline
    \end{tabular}
  \end{adjustbox}
  \caption*{\centering \textbf{TABLE I:} F1-Score Performance Across Configurations: Base, Fine-Tuned on Real Data (20K), and Fine-Tuned on Real + Synthetic Data (100K)}
\end{table*}

\section{Results}
\label{results}

This section evaluates the model's performance before and after synthetic data augmentation. We investigate how increasing the training corpus from 20,000 to 100,000 Arabic medical records via synthetic generation affects the fluency, contextual correctness, and semantic relevance of the generated text.

\subsection{Quantitative Evaluation}

To measure the performance of each model across configurations, we report the BERTScore F1-score as our primary metric. BERTScore offers a semantic evaluation of generated responses by comparing contextual embeddings rather than relying on exact word matches, making it suitable for evaluating generative outputs in NLP.

Table I presents a three-way comparison:

\begin{enumerate}
    \item \textbf{Base Model:} Performance of the pre-trained model without any domain-specific fine-tuning.
    \item \textbf{Fine-Tuned on 20K (Real Only):} Model performance after being trained on the original curated dataset of 20,000 real Arabic patient–doctor interactions.
    \item \textbf{Fine-Tuned on 100K (Real + Synthetic):} Performance after training on the expanded dataset incorporating 80,000 high-quality synthetic samples.
\end{enumerate}

\textbf{Key Observations:}
\begin{itemize}
    \item All models benefited from fine-tuning with both 20K and 100K datasets.
    \item Synthetic augmentation led to a clear increase in F1-score for every model—improving up to +13\% over the 20K fine-tuned version in the case of Mistral-7B.
    \item The best-performing model remained \textbf{Mistral-7B-Instruct-v0.2}, which reached an F1-score of \textbf{81.36\%} after training on the 100K records.
    \item Smaller models like AraGPT2-Base also showed large relative gains (+17\% over base), confirming the utility of synthetic data even in resource-constrained architectures.
\end{itemize}

To further understand the individual contribution of each synthetic data source, we did an ablation study by training models separately on 40,000 records generated by ChatGPT-4o and Gemini 2.5 Pro. The results, discussed in Section~\ref{ablation}, highlight that ChatGPT-4o-generated data consistently produced better model performance and fewer hallucinations than Gemini, reinforcing the importance of data source quality in synthetic augmentation workflows.

\subsection{Qualitative Observations}

In addition to quantitative scores, qualitative trends suggest strong enhancements in generative behavior following synthetic augmentation:

\begin{itemize}
    \item \textbf{Improved Linguistic Fluency:} Responses from synthetic-augmented models exhibited smoother grammar and more natural dialectal phrasing.
    \item \textbf{Broader Vocabulary:} Exposure to synthetic prompts helped models diversify symptom descriptions and adapt to edge cases.
    \item \textbf{Reduced Overfitting:} Compared to the 20K-only setup, responses were less likely to mirror training examples verbatim—indicating better generalization.
    \item \textbf{Stronger Medical Reasoning:} Particularly in Mistral and LLaMA, generated doctor responses became more detailed and logically structured.
\end{itemize}.

\begin{table*}[t]
  \centering

  \begin{tabular}{|l|c|c|c|}
    \hline
    \textbf{Model Name} & \textbf{Base Model} & \textbf{ChatGPT-4o (40K)} & \textbf{Gemini 2.5 Pro (40K)} \\
    \hline
    Meta-LLaMA-2-7B              & 64.00\% & 68.65\% & 65.19\% \\
    AraGPT2-Base (148M)          & 55.45\% & 59.67\% & 56.74\% \\
    OpenAI GPT-2 Medium (355M)   & 53.61\% & 57.13\% & 54.14\% \\
    Mistral-7B-Instruct-v0.2     & 65.77\% & 69.30\% & 67.22\% \\
    BigScience/BLOOM-560M        & 50.12\% & 58.30\% & 52.30\% \\
    \hline
  \end{tabular}
  \caption*{\centering \textbf{TABLE II:}Ablation Study: F1-Score Comparison Between ChatGPT-4o and Gemini 2.5 Pro Synthetic Data (40K Each)}
  \label{tab:ablation-sources}
\end{table*}

\section{Ablation Study}
\label{ablation}

We did an ablation research to better understand how each synthetic data source contributes to overall model performance. Specifically, we investigated how training on synthetic data generated simply from ChatGPT-4o or Gemini 2.5 Pro—each providing 40,000 samples—affected the quality of fine-tuned large language models (LLM). To guarantee a controlled and fair comparison, the training setup and hyperparameters employed in these ablation experiments remained the same as those detailed in the Experimental Design section.

\subsection{Isolating Synthetic Sources}

Table II presents the F1-score results for each model trained in three configurations: 
(1) base model with no fine-tuning, 
(2) fine-tuned using only 40,000 ChatGPT-4o-generated records, 
and (3) fine-tuned using only 40,000 Gemini 2.5 Pro-generated records. 
All models were evaluated using BERTScore F1, consistent with the main experiment setup.

\subsection{Performance Analysis}

Across all evaluated models, synthetic data generated by ChatGPT-4o consistently led to higher F1-scores compared to data generated by Gemini 2.5 Pro. The improvement margin varied across models, but the trend was stable and statistically consistent. For example, Mistral-7B-Instruct-v0.2 improved from a base score of 65.77\% to 69.30\% with ChatGPT-4o data, while only reaching 67.22\% with Gemini data. Similarly, AraGPT2-Base, a relatively lightweight model, saw greater gains when trained on ChatGPT-4o samples (+4.22\%) than with Gemini samples (+1.29\%).

These results indicate that ChatGPT-4o not only provides higher-quality outputs in terms of semantic relevance and linguistic fluency but also enhances downstream LLM performance more effectively.This outcome was corroborated by a qualitative examination of the produced samples. ChatGPT-4o outputs showed less hallucinations, improved adherence to medical context, and more accurate language structure. Gemini 2.5 Pro, while competent, occasionally produced verbose or too generic responses, which may have injected noise into the training process.

\subsection{Implications}

This ablation study highlights the importance of synthetic data quality in the success of domain-specific LLM fine-tuning. Even when trained with the same amount of data and under identical settings, the origin of the synthetic samples significantly influenced the final model performance. These insights support the case for continued refinement of prompt engineering strategies and generative model selection in future data augmentation pipelines.

\section{Conclusion}
\label{conclusion}

This study presents a scalable and effective approach for improving Arabic medical chatbot performance through synthetic data augmentation. Starting from a curated dataset of 20,000 Arabic patient–doctor dialogues collected from social media platforms and medical forums, we expanded the training data to 100,000 entries by generating 80,000 synthetic records using ChatGPT-4o and Gemini 2.5 Pro.

We observed significant gains in semantic accuracy, fluency, and generalization after extensively fine-tuning different LLMs, including Mistral-7B, AraGPT2, and Meta LLaMA-2-7B. Our findings showed that the synthetic data improved overall performance while also reducing overfitting and response repetition. An ablation investigation revealed that ChatGPT-4o-generated data outperformed Gemini in downstream tasks, emphasizing the importance of data source quality in synthetic augmentation pipelines.

Overall, this study establishes synthetic data as a viable and effective option for scaling medical NLP in low-resource languages. By demonstrating its success in Arabic healthcare applications, we lay the groundwork for broader adoption of generative AI in underrepresented language settings.

\bibliographystyle{IEEEtran}

\end{document}